\documentclass[11pt]{article}

\usepackage[final]{acl}

\usepackage{times}
\usepackage{latexsym}

\usepackage[T1]{fontenc}

\usepackage[utf8]{inputenc}

\usepackage{microtype}

\usepackage{inconsolata}

\usepackage{graphicx}

\usepackage{hyperref}
\usepackage{booktabs}
\usepackage{multirow}
\usepackage{multirow}
\usepackage{tabularx}
\usepackage[table,xcdraw]{xcolor} 
\usepackage{amsmath}
\usepackage{amssymb}
\usepackage{mathtools}
\usepackage{amsthm}
\usepackage{bbold}
\usepackage{booktabs}
\usepackage{multirow}
\usepackage{lipsum}
\usepackage{float}
\usepackage{caption} 
\usepackage{color}
\usepackage{listings}
\usepackage[dvipsnames]{xcolor}
\usepackage{makecell}
\usepackage{tcolorbox}
\tcbuselibrary{skins,breakable}
\usepackage{minted}
\usepackage{caption}
\usepackage[utf8]{inputenc} 
\usepackage{multicol}
\usepackage{paracol} 
\usepackage{enumitem}
\lstdefinelanguage{yaml}{
  morekeywords={true,false,null},
  keywordstyle=\color{purple},
  sensitive=false,
  comment=[l]{\#},
  commentstyle=\color{gray},
  stringstyle=\color{teal},
  morestring=[b]',
  morestring=[b]",
  basicstyle=\ttfamily\small,
  breaklines=true
}
\usepackage{booktabs}
\title{"LLM Agent Performance" Is Not a Single Evaluation Target}

\author{
Pengyu Zhu$^{1}$ \qquad
Li Sun$^{1}$ \qquad
Philip S. Yu$^{2}$ \qquad
Sen Su$^{1,3\dagger}$ \\[0.5em]
$^{1}$Beijing University of Posts and Telecommunications \\
$^{2}$University of Illinois Chicago \\
$^{3}$Chongqing University of Posts and Telecommunications \\[0.5em]
\texttt{whfelingyu\_zhupengyu@bupt.edu.cn}
}

\begin{document}
\maketitle
\begin{abstract}
LLM agent benchmark scores are shaped not only by the model but also by the agent harness, environment, evaluator, and inference budget.
Unified execution controls these non-model factors by evaluating candidate models under the same configuration, making observed differences more attributable to the models themselves.
However, model comparison is only one use of agent benchmarks.
Other evaluations compare complete agent systems or test whether a fixed model or system remains stable across predeclared changes in its operating conditions.
These results can all be reported under the common label of "LLM agent performance."
\textbf{Our position is that "LLM agent performance" does not denote a single evaluation target.}
Model comparisons under a reference stack and comparisons of complete agent systems answer different questions, while robustness asks whether either conclusion persists across predeclared conditions.
The claim supported by a score therefore depends on the declared candidate boundary and condition policy.
We derive implications for leaderboards, result reporting, and benchmark versioning, showing how distinguishing these classes preserves fair comparison while accommodating system innovation and robustness analysis.
\end{abstract}
\section{Introduction}
\label{sec:introduction}

LLM agent benchmarks are increasingly used to compare backbone models \cite{agentbench,kapoor2026holistic}, evaluate complete agent systems \cite{bandel2026general}, and assess whether performance persists as interfaces, environments, or resource constraints change.
These purposes are distinct, yet they are often grouped under the label "LLM agent performance"
~\cite{zhang2026harness,yao2026harnessbench,yehudai2025survey}.
Yet a benchmark score does not, by itself, specify which of these questions it answers.

Suppose two LLMs obtain scores of 62 and 58 on the same agent benchmark.
Under a shared execution setup, this difference supports a conditional comparison between the models~\cite{zhu2026unified,kapoor2026holistic}.
If each model is instead embedded in a separately optimized agent, the scores compare complete systems~\cite{bandel2026general}.
If either candidate is evaluated across predeclared changes outside its boundary, the question becomes one of robustness~\cite{yao2026harnessbench,wijk2025rebench}.
Thus, the same benchmark and metric can support different conclusions depending on the declared candidate boundary and condition policy.

This ambiguity is especially consequential for multi-step agents, whose outcomes arise from interactions between a model and an execution stack comprising the harness, inference interface, environment, evaluator, and resource budget. Small differences in this stack can redirect subsequent actions and observations, altering both absolute scores and model rankings~\cite{kapoor2026holistic,zhang2026harness,yao2026harnessbench}. When the model is the candidate, controlled execution addresses this problem by matching non-candidate factors under a declared reference stack, making score differences more credibly attributable to model substitution within that stack. Prior work operationalizes or studies this requirement through fixed scaffolds and controlled environments~\cite{zhu2026unified,chen2025browsecompplus}, protocols for evaluating native agent implementations~\cite{bandel2026general}, and factorial model--harness comparisons~\cite{yao2026harnessbench,kapoor2026holistic}. However, these contributions address execution control, shared infrastructure, or model--harness interactions; they do not by themselves determine the candidate boundary or the comparative claim that a score supports. The same runtime can instantiate controlled model comparisons, complete-system comparisons, or systematic condition sweeps, but these designs license different conclusions.

We argue that "LLM agent performance" is an umbrella term rather than a context-independent quantity.
The comparative attribution supported by a score depends on two design decisions: the candidate boundary, which identifies the entity being compared and which components are candidate-owned, and the condition policy, which specifies whether factors outside that boundary are fixed or semantically matched, or systematically varied.
These choices constrain comparative attribution but do not  determine evaluation validity; task-construct validity, evaluator validity, contamination, and statistical design remain separate requirements. Robustness is cross-cutting: it asks whether either conclusion remains stable under predeclared changes outside the candidate boundary. Because these designs license different claims, their results should not be collapsed into a single ranking.

\section{Why Model-Level Attribution Requires Controlled Execution}
\label{sec:variance}

Controlled execution does not eliminate all variation; it separates intended variation, which defines the evaluation, from incidental variation, which can confound attribution. Differences across benchmark instances in tasks, tools, and domains are often intentional. In a direct model comparison, however, inference, prompting, memory, tool mediation, evaluator behavior, and environment state must be fixed, matched, or sufficiently measured and adjusted for~\cite{kapoor2026holistic,zhu2026unified,zhang2026harness,yao2026harnessbench,chen2025browsecompplus}.
This design supports conditional model comparison under the reference stack, not a harness-independent estimate of capability.

\subsection{Inference Configuration}
\label{sec:inference-configuration}

Inference configuration governs both decoding behavior and the interface through which a model is invoked. These settings are not always standardized across evaluation setups.

\paragraph{Inference Interfaces and Protocols.}
Even with the same model and prompt, provider-specific protocols can affect execution before generation. Content-safety and request-handling policies may filter or reject requests at the interface level. Azure OpenAI, for example, may additionally block requests under its content-management policy~\cite{microsoft2024azurecontent}. When rejection occurs outside model generation, it is not direct evidence of the model's task-solving ability. Yet an answer-only metric may record it as a task failure unless interface events are logged separately. Inference APIs impose different constraints on tool definitions and function schemas, as discussed in Section~\ref{sec:tool-invocation}.

\paragraph{Inference Engine Variability.}
In open-source settings, identical model weights and tokenized inputs can still yield different outputs across inference engines. Such differences have been observed across vLLM~\cite{vllm}, SGLang~\cite{sglang}, and Hugging Face Transformers~\cite{Transformers}, even under nominally matched configurations~\cite{pape2026silenthyperparameterquantifyingimpact}. In a multi-step agent, a small divergence at one decision point can alter the next tool call or observation and propagate through the remaining trajectory \cite{dong2026agentic}.

\begin{figure}[t]
\centering
\includegraphics[width=\linewidth]{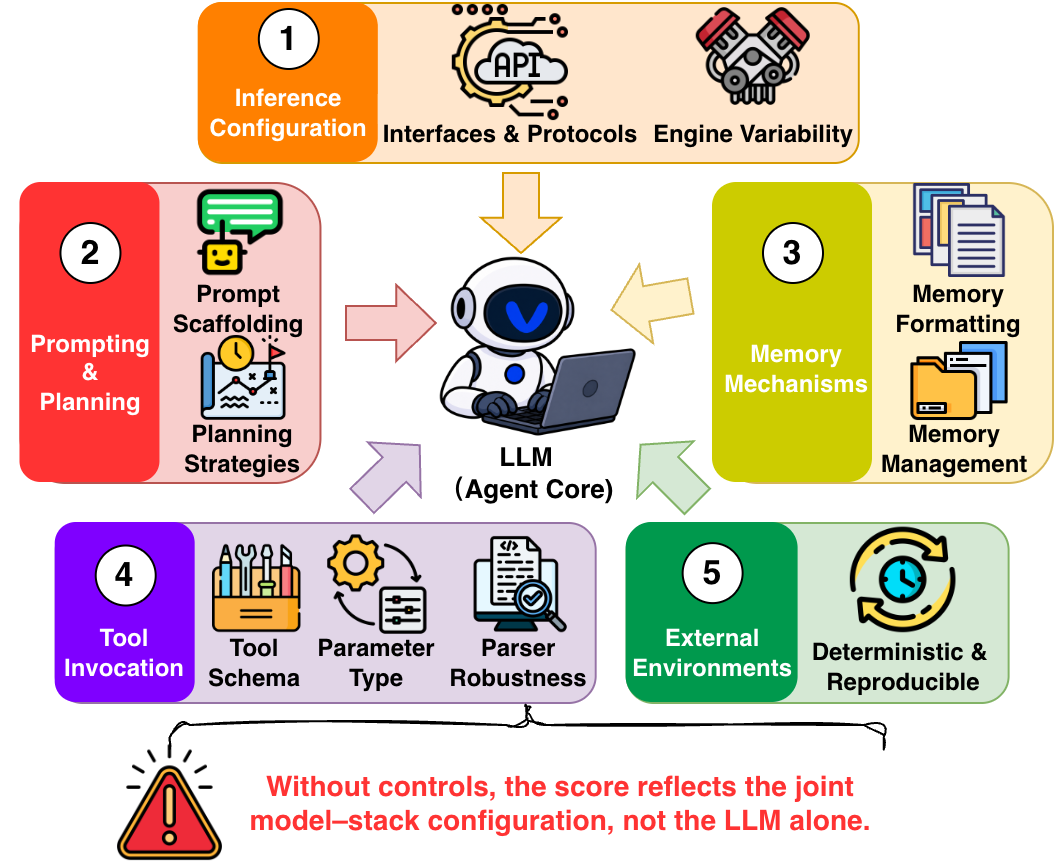}
\caption{Sources of non-model variation that can confound model-level attribution.}
\label{fig:score-targets}
\vspace{-0.5cm}
\end{figure}

\subsection{Prompting and Planning Strategies}

System prompts and planning scaffolds shape how post-trained agentic behavior is elicited during evaluation~\cite{ICLR2024_6c0e99d7,mizrahi-etal-2024-state,dong2026agentic}. Prompts specify tool-use formats and action constraints, while planning scaffolds determine whether an agent decomposes tasks, reflects on errors, or replans. Their role should be specified in the condition policy.

\paragraph{Prompt Scaffolding.}
Benchmark-specific instructions should be distinguished from the procedural scaffold supplied by the execution framework. A function-calling benchmark may prescribe an output format, while a customer-service benchmark may define a role and interaction constraints; these requirements belong to the benchmark specification. By contrast, demonstrations, planning routines, action--observation protocols, retry policies, and recovery instructions are choices made by the scaffold. BFCL and smolagents illustrate this distinction. In the implementations examined, BFCL chiefly specifies benchmark-specific invocation requirements, whereas smolagents embeds its prompt within a broader agent-control policy that includes worked examples, planning, and replanning~\cite{BFCL,smolagents,zhu2026unified}. This contrast does not establish that one benchmark is more difficult or that one scaffold is more effective. Instead, it shows that model outcomes are conditional on how the surrounding scaffold elicits and supports behavior. When the LLM is the candidate, the scaffold should therefore be fixed or semantically matched across models; when complete agent systems are compared, it may legitimately remain candidate-owned.

\paragraph{Planning Strategies.}
Agent harnesses organize reasoning and action through approaches such as Chain-of-Thought prompting~\cite{NEURIPS2022_9d560961}, ReAct~\cite{yao2023react}, or Plan-and-Execute~\cite{wang-etal-2023-plan}. These approaches differ in whether reasoning precedes action, is interleaved with action, or follows an explicit plan. Planning granularity, action constraints, and reflection policies can further alter trajectories within the same approach~\cite{NEURIPS2023_1b44b878}. Improvements obtained by changing this structure are legitimate agent-system advances, but they do not by themselves establish greater capability of the backbone LLM. Controlled model comparison therefore requires a shared planning scaffold; separately optimizing the strategy for each candidate instead produces an agent-system comparison.

\subsection{Memory Mechanisms}

Most explicit agent memory is harness-managed: the harness determines how interaction history is represented, retained, retrieved, and presented, while the model determines how it uses the resulting context~\cite{packer2024memgptllmsoperatingsystems,DBLP:journals/chinaf/XiCGHDHZWJZZFWXZWJZLYDW25}. 

\paragraph{Memory Formatting.}
Harnesses serialize trajectories, tool calls, and outcomes in different ways, thereby changing the effective context presented to the model. One harness may expose observations, actions, tool results, and errors as separate structured fields, while another may flatten them into a chat history. Implementations such as smolagents, $\tau$-bench, and BFCL instantiate different representation and feedback policies~\cite{smolagents,yao2025taubench,BFCL}. The same model can consequently receive materially different inputs across harnesses. In model comparison, unless the serialization policy is fixed or semantically matched, failures and score differences cannot be attributed to the model alone.

\paragraph{Memory Management Under Length Constraints.}
When interaction histories exceed the context window, harnesses must decide what to retain or discard. Strategies range from FIFO truncation to summarization-based compression~\cite{Human-AIinteractionagents} and retrieval- or tiered-memory mechanisms such as MemGPT~\cite{packer2024memgptllmsoperatingsystems}. Harnesses must also determine whether memory persists across episodes. A model may succeed when key observations are retained or retrieved but fail after they have been truncated ~\cite{liu-etal-2024-lost}. Better memory management is a legitimate agent-system advantage. For model-level attribution, candidates must share a memory policy; otherwise, long-horizon scores characterize model--memory configurations, not LLMs alone.

\subsection{Tool Invocation}
\label{sec:tool-invocation}

Tool invocation depends on how tools are represented, which parameter types are exposed, and how model outputs are converted into executable calls. These choices determine both which actions a model can express and which attempted calls are accepted. They should therefore be specified and controlled in model-level comparisons.

\paragraph{Schema and Interface Compatibility.}
Even when the underlying tool is fixed, the same logical interface may require different declarations across deployment backends. Under OpenAI's strict function-calling mode, object schemas must set \texttt{additionalProperties} to \texttt{false} and mark all properties as required; optional values must instead admit \texttt{null}~\cite{openai2024functioncalling}. Gemini supports only a subset of the OpenAPI schema and imposes additional restrictions on supported parameter types~\cite{google2024functioncalling}. vLLM provides configurable validation behavior: named and required tool calls can use schema-constrained decoding, whereas automatic tool calling may depend on the parser and validation settings~\cite{vllm}.
Open-ended annotations such as \texttt{dict[str, Any]} make these incompatibilities concrete. Representing them across backends may require either narrowing the schema or maintaining backend-specific definitions. In direct model comparisons, translation rules, validation settings, and parser versions must be fixed and disclosed; if candidates optimize these interfaces separately, the resulting comparison is between complete systems.

\paragraph{Parsing and Recovery Policies.}
Evaluation outcomes also depend on how model outputs are translated into executable tool calls. An agent may receive tool calls as structured objects from a provider API, or its serving and harness layers may extract them from generated tokens. In the implementations examined, vLLM's automatic tool-calling path uses a configured, often model-specific tool-call parser, whereas smolagents can consume API-provided tool calls or parse them when they are not returned as structured objects~\cite{vllm,smolagents}. Parsing policy determines which attempted calls are recognized as valid, while recovery policy determines whether invalid calls are surfaced, repaired, or retried. The same attempted action may therefore be accepted under one configuration but rejected or retried under another. When these policies fall inside the declared candidate boundary, their effects are legitimately part of complete-system performance. In controlled model comparison, however, the parsing interface and repair and retry policies should be fixed or semantically matched across models.

\subsection{External Environments}
\label{sec:external-environments}

The external environment determines the evidence and state transitions available to an agent. Live search results, source availability, and tool responses may vary across runs, so an incorrect outcome cannot by itself be attributed to the model's reasoning or capability. Controlled model comparison therefore requires candidates to face fixed, replayable, or otherwise matched environment conditions~\cite{ICLR2024_4410c071}. Agent-system evaluations may retain live conditions, and robustness evaluations may vary them deliberately, but the resulting claims remain conditional on those conditions.

BrowseComp~\cite{wei2025browsecompsimplechallengingbenchmark} does not provide a fixed retrieval corpus, so browsing-enabled evaluations depend on the search service and web state at evaluation time. BrowseComp-Plus~\cite{chen2025browsecompplus} provides a fixed corpus with verified documents, enabling more controlled attribution. BFCL~V4 standardizes the search and webpage-fetch interfaces, but the returned evidence can still vary~\cite{BFCL}. These examples show that a shared interface alone does not eliminate environmental variation. Outcome-only scores provide insufficient evidence for distinguishing model failures from environmental failures~\cite{NEURIPS2024_877b4068}; such attribution requires logged observations and controlled comparisons~\cite{zhu2026unified}. 
\section{Control Is Necessary, but Does Not Determine the Evaluation Target}
\label{sec:necessity}

The control-of-variables principle underpins controlled model comparison.
Established benchmarking practice goes beyond holding conditions constant: it defines the candidate boundary, records the reference configuration, and avoids treating results obtained under changed conditions as directly comparable.
Three familiar analogies clarify complementary aspects of this argument.

\subsection{Three Analogies, Three Lessons}
\label{sec:analogy}

\paragraph{Wind tunnels: capability is conditional.}
To compare two airfoils, engineers test them under matched flow and mounting conditions.
The resulting aerodynamic coefficients are interpretable because those conditions are controlled, but they remain measurements \emph{under those conditions}, not guarantees under untested operating conditions.
Likewise, a fixed agent harness makes the LLM's contribution measurable, but the result remains conditional on that reference harness and environment.

\paragraph{GPU and system benchmarks: provenance defines comparability.}
A GPU benchmark score is produced by a broader hardware--software stack. If the goal is to isolate GPU performance, the CPU, memory, driver, and workload must be matched across candidates. Benchmark suites such as MLPerf and 3DMark also record the tested system, workload, software stack, and benchmark version~\cite{reddi2020mlperf,3dmark}. When these conditions differ, scores should not be assumed directly comparable or attributed to the GPU. Likewise, model-level agent comparison requires matched non-model conditions, while every result should include the harness and environment versions that define its comparison class.

\paragraph{Motorsport: component and system competitions are different.}
An engine test bench can compare power units under prescribed loads; a Formula~1 lap time reflects the car, driver, tires, and strategy operating under a common regulatory regime.
Both are valid measurements, but a faster lap cannot be attributed to the engine alone.
Technical regulations define a comparison class without turning a full-car result into a component result.
Agent evaluation needs separate regimes for an LLM under a reference stack and for an optimized end-to-end agent.
Harness innovation should count in the latter and be controlled in the former.

\subsection{What Controlled Execution Establishes and What It Does Not}

The three analogies point to a common conclusion: control makes a comparison interpretable, but it does not determine the candidate being compared. Unified execution addresses one specific attribution problem. When the goal is to compare LLMs, candidates should face the same harness, environment, evaluator, and resource budget. Harness studies show why such control matters~\cite{kapoor2026holistic}, while unified and interoperable frameworks demonstrate that it can be implemented across heterogeneous benchmarks~\cite{zhu2026unified,bandel2026general,Harbor_Framework}.

A shared runtime, however, does not determine what the evaluation measures. If the harness is fixed and only the LLM varies, the resulting score supports a model-level comparison. If participants may supply or optimize the complete agent, the score supports an agent-system comparison. If a declared model or system is tested across predeclared changes in execution conditions, the evaluation assesses robustness. These designs may use the same API or sandbox, but they answer different questions and their scores are not interchangeable.

The evaluation question must therefore be specified before the controls are chosen. Scores from these designs should not be collapsed into a single ranking. Before reporting an agent score, an evaluation should answer four questions:
\begin{enumerate}[leftmargin=*,nosep]
\item What is the candidate: an LLM evaluated through a reference stack or a complete executable agent system?
\item Which components are fixed, candidate-owned, or systematically varied?
\item Is the result a reference-condition comparison or a robustness claim across declared conditions?
\item Which other results were obtained under compatible targets and conditions?
\end{enumerate}

\section{Candidate Boundaries and Condition Policies Behind
"LLM Agent Performance"}
\label{sec:targets}

\begin{figure*}[t]
\centering
\includegraphics[width=\textwidth]{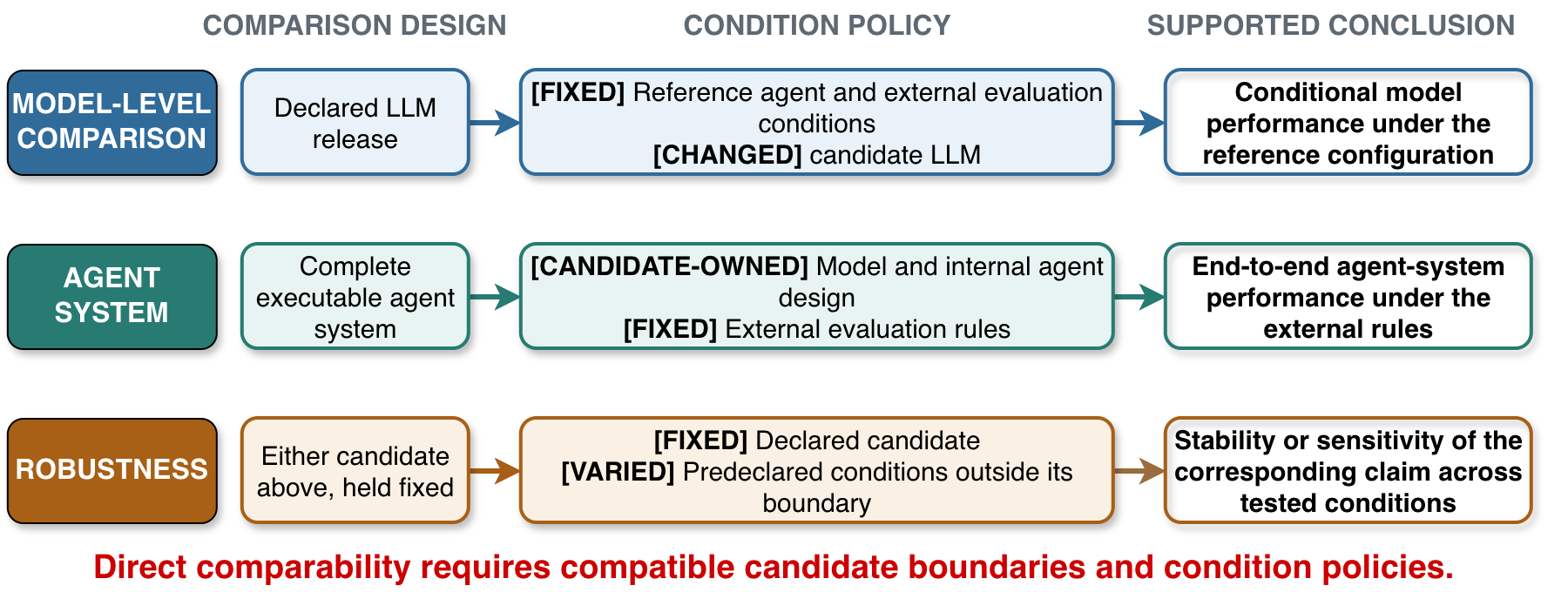}
\caption{Candidate boundaries and condition policies determine which claims agent evaluations support. Robustness is cross-cutting: it tests either candidate across predeclared conditions outside its boundary.}
\label{fig:three-targets}
\vspace{-0.5cm}
\end{figure*}

The phrase "LLM agent performance" conflates two design decisions.
The candidate boundary identifies the compared entity and its
candidate-owned components.
The condition policy specifies whether factors outside that boundary are
held at reference values, semantically matched across candidates, or
systematically varied over a predeclared set.
A material difference that is neither candidate-owned nor included in the declared condition sweep is an uncontrolled confound. Combining two canonical candidate boundaries with two condition policies yields four comparison classes. Robustness is cross-cutting: it asks whether either type of conclusion remains stable across specified changes.
Figure~\ref{fig:three-targets} summarizes this distinction.

\subsection{Comparing Models under Shared Execution}
A model-level evaluation asks which declared LLM release performs
better within the same reference agent.
The model is the intended source of variation, while prompting, planning,
memory, tool mediation, environment, evaluator, and inference budget are
held fixed across candidates.
Unified frameworks can support this target when they hold the
scaffold and execution conditions fixed across
models~\cite{zhu2026unified,kapoor2026holistic}.
The resulting claim is conditional: model \(m_1\) outperforms model
\(m_2\) under the declared reference configuration.
Thus, this design supports model attribution within that configuration; it does not characterize separately optimized agent systems.

\subsection{Comparing Complete Agent Systems}

An agent-system evaluation asks which complete agent system performs
better under common external evaluation rules.
The candidate includes the backbone model, inference configuration,
prompts, planning and memory mechanisms, tool routing and parsing, and
other candidate-owned execution choices.
The evaluation must declare which tools and external services belong
to the candidate and apply common evaluation rules for tasks,
environments, evaluators, safety, and resource use.
General Agent Evaluation exemplifies this target by comparing
general-agent implementations across multiple
environments~\cite{bandel2026general}.
Infrastructure such as Harbor and AgentCompass can instantiate this
design when the evaluated candidate includes both the model and its
harness, rather than only the LLM within a fixed
scaffold~\cite{Harbor_Framework,chen2026agentcompassunifiedevaluationinfrastructure}.
The resulting claim is conditional: one declared agent system
outperforms another under those external rules.
The score does not isolate the contribution of the LLM.

\subsection{Testing Robustness across Declared Conditions}

A robustness evaluation fixes a declared candidate, either an LLM under a reference agent or a complete agent system, and tests it across a predeclared condition sweep~\cite{rabanser2026scienceaiagentreliability}.
BrowseComp and BrowseComp-Plus illustrate why environmental conditions
matter: live-web evaluation depends on changing search results and
content availability, whereas a fixed corpus provides a reproducible
reference environment~\cite{wei2025browsecompsimplechallengingbenchmark,chen2025browsecompplus}.
However, evaluation on a fixed corpus alone does not measure robustness.
A robustness study instead evaluates the same candidate under a
declared set of conditions, such as live-web access, a fixed snapshot,
multiple time-indexed snapshots, alternative reference harnesses when the candidate is an LLM,
or different external budgets.
Robustness therefore qualifies a model-level or agent-system claim
by identifying the tested conditions over which it remains stable; it
does not produce a context-free performance score.

\subsection{The Evaluation Target Determines What Must Be Fixed}

A component's role depends on what the study treats as the candidate. If the candidate is the LLM, prompts, planning, memory, routing, and parsing are part of the evaluation setup and should be fixed or matched across models. If the candidate is the complete agent system, these components are part of the candidate and may differ across systems. Robustness evaluation follows a separate rule: the candidate remains fixed while declared conditions outside its boundary are varied. Thus, changing the prompt may test the robustness of a fixed LLM, but it does not test the robustness of a fixed agent system whose prompt is an internal component. Any systematic cross-run difference must be part of the declared candidate or a deliberately varied condition; otherwise, it confounds the comparison.

\section{Implications for Comparing and Reporting Agent Scores}
\label{sec:comparison}

Once the evaluation target is explicit, direct comparability requires a shared candidate-boundary rule and compatible condition policies; a common benchmark name is insufficient.
Before placing scores in the same comparison set, a leaderboard should declare what the candidate includes, whether external conditions are matched or systematically varied, and which task, evaluator, and resource specification applies.

\subsection{When Scores Are Comparable}

Two scores are directly comparable only when they address the same
evaluation target, use the same candidate-boundary rule, and follow
compatible condition policies.
What counts as compatible depends on the target.
Model-level comparisons should share non-model execution
conditions.
Agent-system comparisons may vary candidate-owned components under
common external rules.
Robustness comparisons require the same declared condition set, or
condition sets whose relationship is explicitly analyzed.
Compatibility therefore does not require every component to be
identical across all evaluation designs; it requires that any unmatched
factor be permitted by the declared target and condition policy.

Within each condition, evaluations should align the task set, scoring
rule, aggregation procedure, uncertainty treatment, and other factors
that are neither part of the candidate nor deliberately varied.
Stochastic evaluations should report uncertainty estimated from
repeated runs or another justified sampling procedure~\cite{NEURIPS2021_f514cec8}.

Benchmark revisions complicate comparability.
Provider APIs change, environments drift, evaluators are updated, and
harness bugs are corrected.
Appending a post-revision result to an earlier ranking assumes that the
change affects all candidates similarly, an assumption undermined by documented harness sensitivity~\cite{kapoor2026holistic,zhang2026harness,alzahrani-etal-2024-benchmarks}.

When continuity across versions matters, benchmark maintainers should
evaluate common anchor candidates under both configurations, using
paired tasks where possible.
These bridge evaluations can reveal score shifts, rank correlations,
rank reversals, and candidate-specific effects.
If score shifts are approximately uniform and rankings remain stable,
the releases may be presented as a linked but versioned series, with the
bridge evidence reported.
Otherwise, they should be treated as distinct comparison regimes.
A bridge establishes continuity only within the same evaluation target
and candidate boundary; it cannot convert agent-system evidence into a
model-level claim.

\subsection{Separate Targets and Match Claims to Candidates}

Results from different targets may be presented together for diagnosis,
but they should not be merged into a single scalar ranking~\cite{ethayarajh-jurafsky-2020-utility,pmlr-v235-zhang24u}.
A model may lead under a reference harness, while another backbone may
support the strongest optimized system.
The same or another candidate may be more stable across changing
conditions.
These outcomes describe different properties rather than contradictory
measurements.
Leaderboards should maintain separate tracks for model
capability and agent-system performance.
Because robustness is cross-cutting, it should be reported as a
condition-aware layer within each track, or in a panel that preserves the underlying candidate type.

The subject and scope of each reported conclusion should match the
candidate and tested conditions.
Appropriate formulations include:
\begin{itemize}[leftmargin=*,nosep]
\item Under the declared reference configuration, model \(m_1\)
outperformed model \(m_2\).
\item Under the declared external rules, agent system \(a_1\)
outperformed agent system \(a_2\).
\item Across the declared condition set $\mathcal{C}$, candidate
\(c\)'s score ranged from $s_{\min}$ to $s_{\max}$.
\end{itemize}

At minimum, reports should specify:
\begin{itemize}[leftmargin=*,nosep]
\item the evaluation target and candidate boundary;
\item the components that were fixed, candidate-owned, or
deliberately varied;
\item benchmark, task-set, harness, environment, evaluator, and
inference-interface versions;
\item resource budgets, retry policies, scoring rules, and aggregation
procedures; and
\item uncertainty estimates and known comparability limitations.
\end{itemize}

Together, these requirements define a target-first principle: evaluators should declare the candidate boundary and condition policy before aggregating scores. Results should be compared only within the same comparison class, defined by a common evaluation target and compatible conditions, and this scope should be explicit in leaderboard labels and reported claims. Unified infrastructure can support any such design, but it cannot choose the evaluation target: it standardizes how an evaluation is executed, whereas the candidate boundary and condition policy determine what is compared and which claims the resulting scores warrant.

\section{Alternative Views}
\label{sec:alternative}
We address six principled objections to clarify our position's scope and limits.

\paragraph{``Static LLM Benchmarks Already Capture Agentic Capability''}
If agentic capability is a property of an LLM, one might expect static
input--output benchmarks to measure it directly.
Such benchmarks can probe prerequisites for agent behavior, including
instruction following, planning, and tool-selection knowledge; offline
trajectory evaluations can also assess decisions conditioned on supplied
histories~\cite{yehudai2025survey}.
Neither setting, by itself, measures how the candidate's own actions alter
later observations through tool feedback and environment transitions
~\cite{NEURIPS2024_877b4068}.
Claims about realized closed-loop behavior therefore require a declared
interactive setup and controls appropriate to the intended claim.

\paragraph{``Standardization Slows Innovation in Agent Research''}
A reasonable concern is that a common harness may privilege one agent
design and deny credit to improvements in planning, memory, or tools.
Our position does not require a fixed harness for every evaluation target.
In model-level evaluation, the reference stack is fixed or matched to
support a conditional comparison between models.
In agent-system evaluation, internal mechanisms are candidate-owned
and may be optimized under common external rules.
Separate tracks preserve harness innovation without attributing
system-level gains to the backbone model.

\paragraph{``Existing Frameworks Already Make Agent Scores Comparable''}
Recent work already instantiates several of the comparison designs
distinguished here.
\citet{zhu2026unified} execute candidate models through a fixed
ReAct-style architecture in a controllable sandbox, thereby supporting
model comparison under a shared reference stack.
HAL standardizes evaluation infrastructure and analyzes variation across
models, scaffolds, and benchmarks~\cite{kapoor2026holistic}.
In our terminology, its within-scaffold contrasts support conditional
model comparisons, whereas scaffold-varying contrasts concern
model--scaffold configurations.
General Agent Evaluation crosses agent architectures with backbone models,
whereas Harness-Bench evaluates model--harness configurations under shared
external conditions~\cite{bandel2026general,yao2026harnessbench}.
Harbor provides an execution substrate capable of instantiating several
such designs~\cite{Harbor_Framework}, but it does not choose the candidate
boundary.
\citet{zhang2026harness} address harness disclosure and model--harness
attribution through locked-harness and factorial comparisons.
Our position is independent of whether model or harness variance dominates:
candidate boundaries and condition policies determine whether a score
supports a model-level, system-level, or cross-condition claim.

\paragraph{``Production Agents Should Replace Reference Harnesses''}
Production agents are appropriate candidates for end-to-end agent-system
evaluation because the deployed model, instructions, tools, state
management, and recovery logic jointly define the candidate.
The resulting score therefore supports a system-level claim, not isolated
attribution to the backbone model.
Such systems are not automatically appropriate reference harnesses for
cross-model comparison, especially when their internal design is coupled
to a particular model.
Substituting another model may then measure compatibility with that stack
rather than a model-level difference under a shared design.
System-level evaluation answers deployment choice, whereas model-level
evaluation is still needed to compare model substitutions within a fixed
stack and distinguish model gains from agent-engineering gains.
Such comparisons require a declared reference stack applied consistently
across candidates.
The stack need not be neutral, but its configuration and the conditional
scope of the result must be reported.

\paragraph{``MCP and Similar Protocol Efforts Are Sufficient''}
The Model Context Protocol standardizes context exchange and tool,
resource, and prompt exposure while leaving their use to host
applications~\cite{hou2025modelcontextprotocolmcp}.
This improves interoperability but does not, by itself, specify an
evaluation design.
It leaves prompt assembly, memory and recovery policies, environment and
evaluator versions, and resource budgets to the implementation.
Nor does it determine whether the candidate is an LLM or a complete agent
system.
MCP can support multiple evaluation designs, but a shared
protocol does not make results directly comparable across candidate
boundaries and condition policies.

\paragraph{``Sandboxed Environments Sacrifice Ecological Validity''}
A sandbox may omit dynamics present in deployment, but our position does
not require universal determinism.
Model comparison requires candidates to face matched environmental
specifications and external conditions, not identical trajectories.
WebArena shows that a self-hosted web environment can be both realistic
and reproducible, whereas BrowseComp-Plus uses a fixed corpus to reduce
dependence on dynamic and opaque search services
~\cite{ICLR2024_4410c071,chen2025browsecompplus}.
Seeded simulators and paired repeated live runs provide other forms of
control, although they do not support identical claims.
Agent-system comparisons may retain live dynamics, but candidates should
still face a common task distribution, access policy, evaluator, and
resource budget.
Robustness analysis instead varies declared environmental conditions
systematically across a fixed candidate.
The environment policy should match the evaluation target and
be attached to the resulting claim.
Controlled and live evaluations provide complementary evidence rather
than substitutes for one another.


\section{Conclusion}
"LLM agent performance" does not denote a single evaluation target.
Benchmarks may compare LLMs under a reference stack, complete systems
under common rules, or either candidate across declared conditions.
Leaderboards should declare the candidate boundary and condition policy,
separate incompatible comparisons, and report robustness only over
tested conditions.
\section{Limitations}
\label{sec:limitations}
This position isolates how candidate boundaries and condition policies constrain comparative attribution. It does not resolve task-construct validity, evaluator bias, benchmark contamination, or statistical adequacy. A reference harness is not neutral and may interact differently with different model families. Compatibility between condition policies remains domain-dependent, and no finite perturbation set establishes robustness to all deployment settings. Our argument primarily concerns LLM agents; its extension to embodied or safety-focused evaluations requires further study.
\bibliography{custom}

@article{DBLP:journals/chinaf/XiCGHDHZWJZZFWXZWJZLYDW25,
  author={Zhiheng Xi and Wenxiang Chen and Xin Guo and Wei He and Yiwen Ding and Boyang Hong and Ming Zhang and Junzhe Wang and Senjie Jin and Enyu Zhou and Rui Zheng and Xiaoran Fan and Xiao Wang and Limao Xiong and Yuhao Zhou and Weiran Wang and Changhao Jiang and Yicheng Zou and Xiangyang Liu and Zhangyue Yin and Shihan Dou and Rongxiang Weng and Wenjuan Qin and Yongyan Zheng and Xipeng Qiu and Xuanjing Huang and Qi Zhang and Tao Gui},
  title={The rise and potential of large language model based agents: a survey},
  year={2025},
  cdate={1735689600000},
  journal={Sci. China Inf. Sci.},
  volume={68},
  number={2},
  url={https://doi.org/10.1007/s11432-024-4222-0}
}

@Misc{smolagents,
  title =        {`smolagents`: a smol library to build great agentic systems.},
  author =       {Aymeric Roucher and Albert Villanova del Moral and Thomas Wolf and Leandro von Werra and Erik Kaunismäki},
  howpublished = {\url{https://github.com/huggingface/smolagents}},
  year =         {2025}
}

@misc{hou2025modelcontextprotocolmcp,
      title={Model Context Protocol (MCP): Landscape, Security Threats, and Future Research Directions}, 
      author={Xinyi Hou and Yanjie Zhao and Shenao Wang and Haoyu Wang},
      year={2025},
      eprint={2503.23278},
      archivePrefix={arXiv},
      primaryClass={cs.CR},
      url={https://arxiv.org/abs/2503.23278}, 
}

@misc{microsoft2024azurecontent,
  title={Azure {OpenAI} Service content filtering},
  author={Microsoft},
  year={2026},
  howpublished={\url{https://learn.microsoft.com/en-us/azure/ai-services/openai/concepts/content-filter}},
  note={Accessed: 2026-01-16}
}

@misc{openai2024functioncalling,
  title={Function calling - OpenAI API Documentation},
  author={OpenAI},
  year={2026},
  howpublished={\url{https://platform.openai.com/docs/guides/function-calling}},
  note={Accessed: 2026-01-16}
}

@misc{google2024functioncalling,
  title={Function calling | Gemini API | Google AI for Developers},
  author={Google},
  year={2026},
  howpublished={\url{https://ai.google.dev/gemini-api/docs/function-calling}},
  note={Accessed: 2026-01-16}
}

@inproceedings{
yao2025taubench,
title={{$\tau$-bench}: A Benchmark for Tool-Agent-User Interaction in Real-World Domains},
author={Shunyu Yao and Noah Shinn and Pedram Razavi and Karthik R Narasimhan},
booktitle={The Thirteenth International Conference on Learning Representations},
year={2025},
url={https://openreview.net/forum?id=roNSXZpUDN}
}

@inproceedings{vllm,
  title={Efficient Memory Management for Large Language Model Serving with PagedAttention},
  author={Woosuk Kwon and Zhuohan Li and Siyuan Zhuang and Ying Sheng and Lianmin Zheng and Cody Hao Yu and Joseph E. Gonzalez and Hao Zhang and Ion Stoica},
  booktitle={Proceedings of the ACM SIGOPS 29th Symposium on Operating Systems Principles},
  year={2023}
}

@inproceedings{
sglang,
title={{SGL}ang: Efficient Execution of Structured Language Model Programs},
author={Lianmin Zheng and Liangsheng Yin and Zhiqiang Xie and Chuyue Sun and Jeff Huang and Cody Hao Yu and Shiyi Cao and Christos Kozyrakis and Ion Stoica and Joseph E. Gonzalez and Clark Barrett and Ying Sheng},
booktitle={The Thirty-eighth Annual Conference on Neural Information Processing Systems},
year={2024},
url={https://openreview.net/forum?id=VqkAKQibpq}
}

@inproceedings{Transformers,
    title = "Transformers: State-of-the-Art Natural Language Processing",
    author = "Wolf, Thomas  and
      Debut, Lysandre  and
      Sanh, Victor  and
      Chaumond, Julien  and
      Delangue, Clement  and
      Moi, Anthony  and
      Cistac, Pierric  and
      Rault, Tim  and
      Louf, Remi  and
      Funtowicz, Morgan  and
      Davison, Joe  and
      Shleifer, Sam  and
      von Platen, Patrick  and
      Ma, Clara  and
      Jernite, Yacine  and
      Plu, Julien  and
      Xu, Canwen  and
      Le Scao, Teven  and
      Gugger, Sylvain  and
      Drame, Mariama  and
      Lhoest, Quentin  and
      Rush, Alexander",
    editor = "Liu, Qun  and
      Schlangen, David",
    booktitle = "Proceedings of the 2020 Conference on Empirical Methods in Natural Language Processing: System Demonstrations",
    month = oct,
    year = "2020",
    address = "Online",
    publisher = "Association for Computational Linguistics",
    url = "https://aclanthology.org/2020.emnlp-demos.6/",
    doi = "10.18653/v1/2020.emnlp-demos.6
        
        
        
        
        
        
        
        
        
        
        
        
        
        
        
        
        
        
        
        ",
    pages = "38--45"
}

@inproceedings{
yao2023react,
title={ReAct: Synergizing Reasoning and Acting in Language Models},
author={Shunyu Yao and Jeffrey Zhao and Dian Yu and Nan Du and Izhak Shafran and Karthik R Narasimhan and Yuan Cao},
booktitle={The Eleventh International Conference on Learning Representations },
year={2023},
url={https://openreview.net/forum?id=WE_vluYUL-X}
}

@inproceedings{
NEURIPS2022_9d560961,
title={Chain of Thought Prompting Elicits Reasoning in Large Language Models},
author={Jason Wei and Xuezhi Wang and Dale Schuurmans and Maarten Bosma and brian ichter and Fei Xia and Ed H. Chi and Quoc V Le and Denny Zhou},
booktitle={Advances in Neural Information Processing Systems},
editor={Alice H. Oh and Alekh Agarwal and Danielle Belgrave and Kyunghyun Cho},
year={2022},
url={https://openreview.net/forum?id=_VjQlMeSB_J}
}

@inproceedings{wang-etal-2023-plan,
author = {He, Gaole and Demartini, Gianluca and Gadiraju, Ujwal},
title = {Plan-Then-Execute: An Empirical Study of User Trust and Team Performance When Using LLM Agents As A Daily Assistant},
year = {2025},
isbn = {9798400713941},
publisher = {Association for Computing Machinery},
address = {New York, NY, USA},
url = {https://doi.org/10.1145/3706598.3713218},
doi = {10.1145/3706598.3713218},
booktitle = {Proceedings of the 2025 CHI Conference on Human Factors in Computing Systems},
articleno = {414},
numpages = {22},
location = {
},
series = {CHI '25}
}

@misc{packer2024memgptllmsoperatingsystems,
      title={MemGPT: Towards LLMs as Operating Systems}, 
      author={Charles Packer and Sarah Wooders and Kevin Lin and Vivian Fang and Shishir G. Patil and Ion Stoica and Joseph E. Gonzalez},
      year={2024},
      eprint={2310.08560},
      archivePrefix={arXiv},
      primaryClass={cs.AI},
      url={https://arxiv.org/abs/2310.08560}, 
}

@inproceedings{Human-AIinteractionagents,
author = {Park, Joon Sung and O'Brien, Joseph and Cai, Carrie Jun and Morris, Meredith Ringel and Liang, Percy and Bernstein, Michael S.},
title = {Generative Agents: Interactive Simulacra of Human Behavior},
year = {2023},
isbn = {9798400701320},
publisher = {Association for Computing Machinery},
address = {New York, NY, USA},
url = {https://doi.org/10.1145/3586183.3606763},
doi = {10.1145/3586183.3606763},
booktitle = {Proceedings of the 36th Annual ACM Symposium on User Interface Software and Technology},
articleno = {2},
numpages = {22},
location = {San Francisco, CA, USA},
series = {UIST '23}
}

@misc{wei2025browsecompsimplechallengingbenchmark,
      title={BrowseComp: A Simple Yet Challenging Benchmark for Browsing Agents}, 
      author={Jason Wei and Zhiqing Sun and Spencer Papay and Scott McKinney and Jeffrey Han and Isa Fulford and Hyung Won Chung and Alex Tachard Passos and William Fedus and Amelia Glaese},
      year={2025},
      eprint={2504.12516},
      archivePrefix={arXiv},
      primaryClass={cs.CL},
      url={https://arxiv.org/abs/2504.12516}, 
}

@inproceedings{
chen2025browsecompplus,
title={BrowseComp-Plus: A More Fair and Transparent Evaluation  Benchmark of Deep-Research Agent},
author={Zijian Chen and Xueguang Ma and Shengyao Zhuang and Ping Nie and Kai Zou and Sahel Sharifymoghaddam and Andrew Liu and Joshua Green and Kshama Patel and Ruoxi Meng and Mingyi Su and Yanxi Li and Haoran Hong and Xinyu Shi and Xuye Liu and Nandan Thakur and Crystina Zhang and Luyu Gao and Wenhu Chen and Jimmy Lin},
booktitle={First Workshop on Multi-Turn Interactions in Large Language Models},
year={2025},
url={https://openreview.net/forum?id=YJAA2PzfDi}
}

@inproceedings{
BFCL,
title={The Berkeley Function Calling Leaderboard ({BFCL}): From Tool Use to Agentic Evaluation of Large Language Models},
author={Shishir G Patil and Huanzhi Mao and Fanjia Yan and Charlie Cheng-Jie Ji and Vishnu Suresh and Ion Stoica and Joseph E. Gonzalez},
booktitle={Forty-second International Conference on Machine Learning},
year={2025},
url={https://openreview.net/forum?id=2GmDdhBdDk}
}

@inproceedings{agentbench,
title={AgentBench: Evaluating {LLM}s as Agents},
author={Xiao Liu and Hao Yu and Hanchen Zhang and Yifan Xu and Xuanyu Lei and Hanyu Lai and Yu Gu and Hangliang Ding and Kaiwen Men and Kejuan Yang and Shudan Zhang and Xiang Deng and Aohan Zeng and Zhengxiao Du and Chenhui Zhang and Sheng Shen and Tianjun Zhang and Yu Su and Huan Sun and Minlie Huang and Yuxiao Dong and Jie Tang},
booktitle={The Twelfth International Conference on Learning Representations},
year={2024},
url={https://openreview.net/forum?id=zAdUB0aCTQ}
}

@inproceedings{
kapoor2026holistic,
title={Holistic Agent Leaderboard: The Missing Infrastructure for {AI} Agent Evaluation},
author={Sayash Kapoor and Benedikt Stroebl and Peter Kirgis and Nitya Nadgir and Zachary S Siegel and Boyi Wei and Tianci Xue and Ziru Chen and Felix Chen and Saiteja Utpala and Franck Ndzomga and Dheeraj Oruganty and Sophie Luskin and Kangheng Liu and Botao Yu and Amit Arora and Dongyoon Hahm and Harsh Trivedi and Huan Sun and Juyong Lee and Tengjun Jin and Yifan Mai and Yifei Zhou and Yuxuan Zhu and Rishi Bommasani and Daniel Kang and Dawn Song and Peter Henderson and Yu Su and Percy Liang and Arvind Narayanan},
booktitle={The Fourteenth International Conference on Learning Representations},
year={2026},
url={https://openreview.net/forum?id=vUaY1t64ZZ}
}

@misc{Harbor_Framework,
author = {{Harbor Framework Team}},
month = jan,
title = {{Harbor: A framework for evaluating and optimizing agents and models in container environments}},
url = {https://github.com/harbor-framework/harbor},
year = {2026}
}

@inproceedings{reddi2020mlperf,
  title={Mlperf inference benchmark},
  author={Reddi, Vijay Janapa and Cheng, Christine and Kanter, David and Mattson, Peter and Schmuelling, Guenther and Wu, Carole-Jean and Anderson, Brian and Breughe, Maximilien and Charlebois, Mark and Chou, William and others},
  booktitle={2020 ACM/IEEE 47th Annual International Symposium on Computer Architecture (ISCA)},
  pages={446--459},
  year={2020},
  organization={IEEE}
}

@misc{3dmark,
  author       = {{UL Solutions}},
  title        = {{3DMark}: The Gamer's Benchmark for {GPU} Performance},
  year         = {2025},
  howpublished = {\url{https://benchmarks.ul.com/3dmark}},
  note         = {Computer benchmarking tool by UL (formerly Futuremark)}
}

@inproceedings{yehudai2025survey,
    title = "A Survey on Evaluation of {LLM}-based Agents",
    author = "Yehudai, Asaf  and
      Eden, Lilach  and
      Li, Alan  and
      Uziel, Guy  and
      Zhao, Yilun  and
      Bar-Haim, Roy  and
      Cohan, Arman  and
      Shmueli-Scheuer, Michal",
    editor = "Liakata, Maria  and
      Moreira, Viviane P.  and
      Zhang, Jiajun  and
      Jurgens, David",
    booktitle = "Findings of the {A}ssociation for {C}omputational {L}inguistics: {ACL} 2026",
    month = jul,
    year = "2026",
    address = "San Diego, California, United States",
    publisher = "Association for Computational Linguistics",
    url = "https://aclanthology.org/2026.findings-acl.1330/",
    doi = "10.18653/v1/2026.findings-acl.1330",
    pages = "26690--26714",
    ISBN = "979-8-89176-395-1"
}

@misc{zhu2026unified,
  title={A Unified Framework for the Evaluation of {LLM} Agentic Capabilities},
  author={Pengyu Zhu and Lijun Li and Yaxing Lyu and Qianxin Luo and Jingyi Yang and Yi Liu and Tingfeng Hui and Xinyu Yuan and Li Sun and Sen Su and Jing Shao},
  year={2026},
  eprint={2605.27898},
  archivePrefix={arXiv},
  primaryClass={cs.AI},
  url={https://arxiv.org/abs/2605.27898}
}

@misc{zhang2026harness,
  title={Stop Comparing {LLM} Agents Without Disclosing the Harness},
  author={Yunbei Zhang and Janet Wang and Yingqiang Ge and Weijie Xu and Jihun Hamm and Chandan K. Reddy},
  year={2026},
  eprint={2605.23950},
  archivePrefix={arXiv},
  primaryClass={cs.AI},
  url={https://arxiv.org/abs/2605.23950}
}

@misc{bandel2026general,
  title={General Agent Evaluation},
  author={Elron Bandel and Asaf Yehudai and Lilach Eden and Yehoshua Sagron and Yotam Perlitz and Elad Venezian and Natalia Razinkov and Natan Ergas and Shlomit Shachor Ifergan and Segev Shlomov and Michal Jacovi and Leshem Choshen and Liat Ein-Dor and Yoav Katz and Michal Shmueli-Scheuer},
  year={2026},
  eprint={2602.22953},
  archivePrefix={arXiv},
  primaryClass={cs.AI},
  url={https://arxiv.org/abs/2602.22953}
}

@misc{yao2026harnessbench,
  title={Harness-Bench: Measuring Harness Effects across Models in Realistic Agent Workflows},
  author={Yilun Yao and Xinyu Tan and Chao-Hsuan Liu and Yaoming Li and Zhengyang Wang and Wenhan Yu and Zhewen Tan and Yuxuan Tian and Guangxiang Zhao and Lin Sun and Xiangzheng Zhang and Tong Yang},
  year={2026},
  eprint={2605.27922},
  archivePrefix={arXiv},
  primaryClass={cs.AI},
  url={https://arxiv.org/abs/2605.27922}
}

@inproceedings{
wijk2025rebench,
title={{RE}-Bench: Evaluating Frontier {AI} R\&D Capabilities of Language Model Agents against Human Experts},
author={Hjalmar Wijk and Tao Roa Lin and Joel Becker and Sami Jawhar and Neev Parikh and Thomas Broadley and Lawrence Chan and Michael Chen and Joshua M Clymer and Jai Dhyani and Elena Ericheva and Katharyn Garcia and Brian Goodrich and Nikola Jurkovic and Megan Kinniment and Aron Lajko and Seraphina Nix and Lucas Jun Koba Sato and William Saunders and Maksym Taran and Ben West and Elizabeth Barnes},
booktitle={Forty-second International Conference on Machine Learning},
year={2025},
url={https://openreview.net/forum?id=3rB0bVU6z6}
}

@misc{chen2026agentcompassunifiedevaluationinfrastructure,
      title={AgentCompass: A Unified Evaluation Infrastructure for Agent Capabilities}, 
      author={Kai Chen and Zichen Ding and Jiaye Ge and Shufan Jiang and Mo Li and Qingqiu Li and Zehao Li and Zonglin Li and Tianhao Liang and Shudong Liu and Zerun Ma and Zixin Shang and Wenhui Tian and Zun Wang and Liwei Wu and Zhenyu Wu and Jun Xu and Bowen Yang and Dingbo Yuan and Qi Zhang and Songyang Zhang and Peiheng Zhou and Dongsheng Zhu},
      year={2026},
      eprint={2607.13705},
      archivePrefix={arXiv},
      primaryClass={cs.AI},
      url={https://arxiv.org/abs/2607.13705}, 
}

@article{mizrahi-etal-2024-state,
    title = "State of What Art? A Call for Multi-Prompt {LLM} Evaluation",
    author = "Mizrahi, Moran  and
      Kaplan, Guy  and
      Malkin, Dan  and
      Dror, Rotem  and
      Shahaf, Dafna  and
      Stanovsky, Gabriel",
    journal = "Transactions of the Association for Computational Linguistics",
    volume = "12",
    year = "2024",
    address = "Cambridge, MA",
    publisher = "MIT Press",
    url = "https://aclanthology.org/2024.tacl-1.52/",
    doi = "10.1162/tacl_a_00681",
    pages = "933--949"
}

@inproceedings{ICLR2024_6c0e99d7,
 author = {Sclar, Melanie and Choi, Yejin and Tsvetkov, Yulia and Suhr, Alane},
 booktitle = {International Conference on Learning Representations},
 editor = {B. Kim and Y. Yue and S. Chaudhuri and K. Fragkiadaki and M. Khan and Y. Sun},
 pages = {25055--25083},
 title = {Quantifying Language Models\textquotesingle  Sensitivity to Spurious Features in Prompt Design or: How I learned to start worrying about prompt formatting},
 url = {https://proceedings.iclr.cc/paper_files/paper/2024/file/6c0e99d736da621403018ca7b32b1a4d-Paper-Conference.pdf},
 volume = {2024},
 year = {2024}
}

@inproceedings{
dong2026agentic,
title={Agentic Reinforced Policy Optimization},
author={Guanting Dong and Hangyu Mao and Kai Ma and Licheng Bao and Yifei Chen and Zhongyuan Wang and Zhongxia Chen and Jiazhen Du and Huiyang Wang and Fuzheng Zhang and Guorui Zhou and Yutao Zhu and Ji-Rong Wen and Zhicheng Dou},
booktitle={The Fourteenth International Conference on Learning Representations},
year={2026},
url={https://openreview.net/forum?id=TX4k7BF6aO}
}

@inproceedings{NEURIPS2023_1b44b878,
 author = {Shinn, Noah and Cassano, Federico and Gopinath, Ashwin and Narasimhan, Karthik and Yao, Shunyu},
 booktitle = {Advances in Neural Information Processing Systems},
 doi = {10.52202/075280-0377},
 editor = {A. Oh and T. Naumann and A. Globerson and K. Saenko and M. Hardt and S. Levine},
 pages = {8634--8652},
 publisher = {Curran Associates, Inc.},
 title = {Reflexion: language agents with verbal reinforcement learning},
 url = {https://proceedings.neurips.cc/paper_files/paper/2023/file/1b44b878bb782e6954cd888628510e90-Paper-Conference.pdf},
 volume = {36},
 year = {2023}
}

@article{liu-etal-2024-lost,
    title = "Lost in the Middle: How Language Models Use Long Contexts",
    author = "Liu, Nelson F.  and
      Lin, Kevin  and
      Hewitt, John  and
      Paranjape, Ashwin  and
      Bevilacqua, Michele  and
      Petroni, Fabio  and
      Liang, Percy",
    journal = "Transactions of the Association for Computational Linguistics",
    volume = "12",
    year = "2024",
    address = "Cambridge, MA",
    publisher = "MIT Press",
    url = "https://aclanthology.org/2024.tacl-1.9/",
    doi = "10.1162/tacl_a_00638",
    pages = "157--173"
}

@inproceedings{ICLR2024_4410c071,
 author = {Zhou, Shuyan and Xu, Frank F and Zhu, Hao and Zhou, Xuhui and Lo, Robert and Sridhar, Abishek and Cheng, Xianyi and Ou, Tianyue and Bisk, Yonatan and Fried, Daniel and Alon, Uri and Neubig, Graham},
 booktitle = {International Conference on Learning Representations},
 editor = {B. Kim and Y. Yue and S. Chaudhuri and K. Fragkiadaki and M. Khan and Y. Sun},
 pages = {15585--15606},
 title = {WebArena: A Realistic Web Environment for Building Autonomous Agents},
 url = {https://proceedings.iclr.cc/paper_files/paper/2024/file/4410c0711e9154a7a2d26f9b3816d1ef-Paper-Conference.pdf},
 volume = {2024},
 year = {2024}
}

@inproceedings{NEURIPS2024_877b4068,
 author = {Ma, Chang and Zhang, Junlei and Zhu, Zhihao and Yang, Cheng and Yang, Yujiu and Jin, Yaohui and Lan, Zhenzhong and Kong, Lingpeng and He, Junxian},
 booktitle = {Advances in Neural Information Processing Systems},
 doi = {10.52202/079017-2365},
 editor = {A. Globerson and L. Mackey and D. Belgrave and A. Fan and U. Paquet and J. Tomczak and C. Zhang},
 pages = {74325--74362},
 publisher = {Curran Associates, Inc.},
 title = {AgentBoard: An Analytical Evaluation Board of Multi-turn LLM Agents},
 url = {https://proceedings.neurips.cc/paper_files/paper/2024/file/877b40688e330a0e2a3fc24084208dfa-Paper-Datasets_and_Benchmarks_Track.pdf},
 volume = {37},
 year = {2024}
}

@inproceedings{NEURIPS2021_f514cec8,
 author = {Agarwal, Rishabh and Schwarzer, Max and Castro, Pablo Samuel and Courville, Aaron and Bellemare, Marc},
 booktitle = {Advances in Neural Information Processing Systems},
 editor = {M. Ranzato and A. Beygelzimer and Y. Dauphin and P.S. Liang and J. Wortman Vaughan},
 pages = {29304--29320},
 publisher = {Curran Associates, Inc.},
 title = {Deep Reinforcement Learning at the Edge of the Statistical Precipice},
 url = {https://proceedings.neurips.cc/paper_files/paper/2021/file/f514cec81cb148559cf475e7426eed5e-Paper.pdf},
 volume = {34},
 year = {2021}
}

@inproceedings{alzahrani-etal-2024-benchmarks,
    title = "When Benchmarks are Targets: Revealing the Sensitivity of Large Language Model Leaderboards",
    author = "Alzahrani, Norah  and
      Alyahya, Hisham  and
      Alnumay, Yazeed  and
      AlRashed, Sultan  and
      Alsubaie, Shaykhah  and
      Almushayqih, Yousef  and
      Mirza, Faisal  and
      Alotaibi, Nouf  and
      Al-Twairesh, Nora  and
      Alowisheq, Areeb  and
      Bari, M Saiful  and
      Khan, Haidar",
    editor = "Ku, Lun-Wei  and
      Martins, Andre  and
      Srikumar, Vivek",
    booktitle = "Proceedings of the 62nd Annual Meeting of the Association for Computational Linguistics (Volume 1: Long Papers)",
    month = aug,
    year = "2024",
    address = "Bangkok, Thailand",
    publisher = "Association for Computational Linguistics",
    url = "https://aclanthology.org/2024.acl-long.744/",
    doi = "10.18653/v1/2024.acl-long.744",
    pages = "13787--13805"
}

@inproceedings{ethayarajh-jurafsky-2020-utility,
    title = "Utility is in the Eye of the User: A Critique of {NLP} Leaderboards",
    author = "Ethayarajh, Kawin  and
      Jurafsky, Dan",
    editor = "Webber, Bonnie  and
      Cohn, Trevor  and
      He, Yulan  and
      Liu, Yang",
    booktitle = "Proceedings of the 2020 Conference on Empirical Methods in Natural Language Processing (EMNLP)",
    month = nov,
    year = "2020",
    address = "Online",
    publisher = "Association for Computational Linguistics",
    url = "https://aclanthology.org/2020.emnlp-main.393/",
    doi = "10.18653/v1/2020.emnlp-main.393",
    pages = "4846--4853"
}

@InProceedings{pmlr-v235-zhang24u,
  title = 	 {Inherent Trade-Offs between Diversity and Stability in Multi-Task Benchmarks},
  author =       {Zhang, Guanhua and Hardt, Moritz},
  booktitle = 	 {Proceedings of the 41st International Conference on Machine Learning},
  pages = 	 {58984--59002},
  year = 	 {2024},
  editor = 	 {Salakhutdinov, Ruslan and Kolter, Zico and Heller, Katherine and Weller, Adrian and Oliver, Nuria and Scarlett, Jonathan and Berkenkamp, Felix},
  volume = 	 {235},
  series = 	 {Proceedings of Machine Learning Research},
  month = 	 {21--27 Jul},
  publisher =    {PMLR},
  url = 	 {https://proceedings.mlr.press/v235/zhang24u.html}
}

@misc{rabanser2026scienceaiagentreliability,
      title={Towards a Science of AI Agent Reliability}, 
      author={Stephan Rabanser and Sayash Kapoor and Peter Kirgis and Kangheng Liu and Saiteja Utpala and Arvind Narayanan},
      year={2026},
      eprint={2602.16666},
      archivePrefix={arXiv},
      primaryClass={cs.AI},
      url={https://arxiv.org/abs/2602.16666}, 
}

@misc{pape2026silenthyperparameterquantifyingimpact,
      title={The Silent Hyperparameter: Quantifying the Impact of Inference Backends on LLM Reproducibility}, 
      author={David Pape and Jonathan Evertz and Lea Schönherr},
      year={2026},
      eprint={2605.19537},
      archivePrefix={arXiv},
      primaryClass={cs.LG},
      url={https://arxiv.org/abs/2605.19537}, 
}
\clearpage
\appendix
\newpage
\section{Prompt Artifacts Across Candidate Boundaries}
\label{app:prompt-comparison}

\begin{table*}[t]
\centering
\small
\begin{tabular}{@{}p{0.22\textwidth}p{0.13\textwidth}p{0.27\textwidth}p{0.27\textwidth}@{}}
\toprule
Evaluation question
& Candidate
& Role of procedural scaffold
& Licensed claim \\
\midrule

Which LLM performs better under a reference stack?
& LLM
& Fixed or semantically matched outside the candidate
& Conditional model difference under the declared stack \\

Which complete agent performs better?
& Executable agent system
& Candidate-owned and permitted to differ
& End-to-end system difference under common external rules \\

How sensitive is an LLM to scaffolding?
& LLM
& Systematically varied as a declared external condition
& Model sensitivity or stability across the tested scaffold set \\

How robust is a complete agent system?
& Executable agent system
& Internal, candidate-owned, and therefore fixed
& System robustness across declared external conditions \\
\bottomrule
\end{tabular}

\caption{Prompt-scaffold roles under different candidate boundaries.
The benchmark contract remains external in each comparison.}
\label{tab:prompt-roles}
\end{table*}
Prompt-side text can play at least two evaluation roles.
Benchmark-specific instructions define the task and its admissible output contract.
Demonstrations, planning rules, and control logic instead form the procedural scaffold.
The scaffold's role depends on the declared candidate boundary.

BFCL and smolagents provide source-level illustrations of these roles, not alternative prompts for a shared task suite.
BFCL supplies benchmark-specific function-calling templates, whereas smolagents supplies configurable agent templates~\cite{BFCL,smolagents}.
We therefore do not compare their scores, prompt lengths, or benchmark difficulty.
\citet{zhu2026unified} report detailed implementation and trajectory evidence concerning scaffold effects.
Here we classify prompt artifacts by evaluation role rather than reproduce that case study.

\paragraph{Selected source anchors.}
Two short excerpts make the distinction concrete:

\begin{quote}\small\ttfamily
\textbf{BFCL:}
``You should only return the function calls in your response.''~\cite{BFCL}

\textbf{smolagents:}
``ALWAYS provide a tool call, else you will fail.''~\cite{smolagents}
\end{quote}

\noindent
These are selected excerpts rather than complete prompts, and their line breaks have been normalized.
The BFCL excerpt illustrates an output contract.
The smolagents excerpt illustrates a procedural control rule.
Neither excerpt demonstrates that one design is superior or causes a particular score change.

The excerpts come from BFCL's classic no-tag tool-call template and the
smolagents \texttt{ToolCallingAgent} template, respectively.
Because both artifacts are versioned and assembled dynamically, reports
should record the exact repository commit, template path, invocation mode,
and enabled options.

\paragraph{Interpretive limit.}
The excerpts support a classification claim, not a causal scaffold-effect claim.
The latter would require deliberately varying the scaffold while matching tasks, model versions, inference interfaces, tools, environments, evaluators, and resource budgets.
The relevant question is therefore not which prompt is longer, but who owns the procedural scaffold under the declared evaluation target.

\section{Memory Policy Across Candidate Boundaries}
\label{app:memory-comparison}

Agent memory should be treated as a policy governing the model-visible interaction history, rather than as a particular internal data structure.
A framework may store typed execution records while presenting only a flattened message sequence to the model.
Conversely, a simple internal log may still generate explicit feedback or summaries during prompt assembly.
Internal fields are evaluation-relevant only when they affect the context, actions, or resources available to the candidate.

Implementations such as smolagents, $\tau$-bench, and BFCL instantiate different trajectory-representation and feedback policies~\cite{smolagents,yao2025taubench,BFCL}.
These implementations illustrate a design range, not a controlled comparison of memory quality or benchmark difficulty.
Implementation-level contrasts and trajectory evidence are reported by ~\citet{zhu2026unified}.
Here we address the distinct question of how memory policy should be assigned under different evaluation targets.

\paragraph{Dimensions of memory policy.}
At least four dimensions should be declared:

\begin{itemize}\itemsep0pt
\item \textbf{Serialization:} how actions, observations, tool results, errors, and intermediate state are encoded in the model-visible context.
\item \textbf{Retention and retrieval:} which past items are retained, truncated, summarized, or retrieved when the available history exceeds the context or storage budget.
\item \textbf{Persistence:} whether memory is reset between tasks and episodes or may carry information across them.
\item \textbf{Feedback and recovery signaling:} how failed actions are exposed to the model and whether retry, repair, or strategy-change instructions are supplied.
\end{itemize}

These dimensions should be distinguished from logging-only metadata.
Timing records, token counts, internal object types, and diagnostic annotations do not affect model behavior unless the harness exposes or acts upon them.

\paragraph{Role under model comparison.}
When the LLM is the candidate, memory policy belongs to the reference stack.
Serialization, retention, retrieval, persistence, and recovery signaling should therefore be fixed or semantically matched across candidate models.
The associated context, storage, and retry budgets should also be held constant.
The resulting score supports a model comparison under that declared policy, not a memory-independent claim about the LLM.

\paragraph{Role under agent-system comparison.}
When the complete executable agent is the candidate, its memory implementation and update policy are candidate-owned.
Systems may legitimately differ in how they store, compress, retrieve, or present prior interactions.
Advantages arising from these choices are system-level advantages and should not be attributed to the backbone model alone.

\paragraph{Role under robustness analysis.}
Robustness analysis first fixes the candidate.
For an LLM candidate, evaluators may vary external memory policies systematically and report sensitivity across the declared condition set.
For a complete agent system, the internal memory module remains part of the candidate.
Replacing that module defines a different system rather than another condition for the same system.
External constraints, such as context limits, storage capacity, episode length, or observation noise, may still be varied.

\paragraph{Reporting and claim scope.}
Reports should identify the memory-policy source and version, the model-visible message assembly, retention and retrieval rules, reset behavior, recovery signaling, and applicable resource limits.
Comparisons should be based on the realized model-visible context and update policy, not solely on internal class definitions.
These disclosures do not make heterogeneous scores directly comparable, but they reveal whether memory is fixed, candidate-owned, or systematically varied.
A causal claim about memory-policy effects would additionally require a controlled intervention that matches the remaining evaluation stack.

\end{document}